\documentclass[twoside,11pt]{article}
\usepackage{pgm,fourier,tikz,bm,subfigure,booktabs,tikz,pgfplots,algorithm,algorithmic,amsmath,blkarray,todonotes,xparse,soul,comment,url,hyperref}
\usetikzlibrary{arrows,plotmarks,decorations.markings,trees,shapes,pgfplots.groupplots}
\tikzset{dot/.style = {circle, fill, minimum size=#1,inner sep=0pt, outer sep=0pt, fill, circle},dot/.default = 6pt}
\tikzset{dot2/.style = {circle, fill, color=black!40,minimum size=6pt,inner sep=0pt, outer sep=0pt, fill, circle}}
\tikzstyle{a}=[->,>=stealth,dashed]
\tikzstyle{a2}=[->,>=stealth]
\newcommand{\vu}[2]{u_#1^{(#2)}}
\newcommand{\vx}[2]{x_#1^{(#2)}}
\newtheorem{myremark}{Remark}
\hypersetup{colorlinks=false,linkcolor=blue,citecolor=magenta}
\newif\ifarxiv
\arxivtrue
\hypersetup{  
bookmarksnumbered}
\hypersetup{  
breaklinks=false,
pdfborderstyle={/S/U/W 0},
citebordercolor=.235 .702 .443,
urlbordercolor=.255 .412 .882,
linkbordercolor=.804 .149 .149,}
\hypersetup{ 
pdfauthor={},
pdftitle={},
pdfkeywords={}}
\makeatletter
\NewDocumentCommand{\LeftComment}{s m}{%
\Statex\IfBooleanF{#1}{\hspace*{\ALG@thistlm}}\(\triangleright\) #2}
\makeatother
\ShortHeadings{Structural Causal Models Are (Solvable by) Credal Networks}{Zaffalon et al.}
\title{Structural Causal Models Are  (Solvable by) Credal Networks}
\author{\Name{Marco Zaffalon} \Email{zaffalon@idsia.ch}\and
\Name{Alessandro Antonucci} \Email{alessandro@idsia.ch}
\and\Name{Rafael Caba\~nas} \Email{rcabanas@idsia.ch}\\
\addr Istituto Dalle Molle di Studi sull'Intelligenza Artificiale (IDSIA), Lugano, Switzerland}
\begin{document}
\maketitle
\begin{abstract}
A structural causal model is made of endogenous (manifest) and exogenous (latent) variables. We show that endogenous observations induce linear constraints on the probabilities of the exogenous variables. This allows to \emph{exactly} map a causal model into a \emph{credal network}. Causal inferences, such as interventions and counterfactuals, can consequently be obtained by standard algorithms for the updating of credal nets. These natively return sharp values in the identifiable case, while intervals corresponding to the exact bounds are produced for unidentifiable queries. A characterization of the causal models that allow
 the map above to be compactly derived is given, along with a discussion about the scalability for general models. This contribution should be regarded as a systematic approach to represent structural causal models by credal networks and hence to systematically compute causal inferences. A number of demonstrative examples is presented to clarify our methodology. Extensive experiments show that approximate algorithms for credal networks can immediately be used to do causal inference in real-size problems.
\end{abstract}
\maketitle
\section{Introduction}\label{sec:intro}
Since early times, dealing with causality has been---and under many respects, still is---a true challenge for scientists and philosophers \citep{hume}. Currently causality is an emerging direction for data science (e.g., \citeauthor{correa2020calculus}, \citeyear{correa2020calculus}), with a wealth of potential applications in diverse domains such as planning \citep{wilkins2014practical} or NLP \citep{asghar2016automatic}, not to mention fields other than Artificial Intelligence such as Economics, Social Science or Medicine. 

Pearl's structural causal models are a natural formalism for causal inference \citep{pearl2009causality}, in particular for their appealing graphical representation. They are also very general and equivalent to the prominent alternative formalisms proposed to handle causality. However, the peculiar features of causal models may render them not always easy to access to a traditional audience, which is instead familiar with pre-existing graphical tools and related procedures.

In this paper, we focus on Pearl's non-parametric structural causal models with discrete variables, and show that they can be represented by \emph{credal networks} \citep{cozman2000credal}. This is a class of imprecise-probabilistic graphical models originally proposed as tools for sensitivity analysis in Bayesian networks. The representation is exact: this means that every query on the causal model can be reformulated as a query on the credal network, which can then be solved by standard algorithms for the latter. An immediate advantage of this outcome is that causal concepts naturally become more familiar concepts, such as the updating of credal networks. Another, more practical, advantage is that such an outcome allows us to systematically compute causal inference via credal nets, be it interventional queries or the more advanced counterfactual type of inference.

It is well known that causal inference is affected by `identification' problems: an inference is said to be \emph{identifiable} when it can be reduced to a precise probabilistic expression (a number representing a probability or an expectation), which can eventually be computed from data. Otherwise it is called \emph{unidentifiable}: one that cannot be reduced to a number no matter the amount of available data. Most of the causal literature focuses on the characterization of models and queries that are identifiable, the most prominent example being the \emph{do calculus} \citep{pearl1995causal}. Relatively few works consider instead the far bigger arena of unidentifiable problems, which are solved by computing tight bounds on the sought probabilities or expectations \citep{balke1997bounds,kang2012inequality,sachs2020}. The computational results in this paper can be read as contributing in particular to this second direction: in fact, credal nets natively return intervals in the unidentifiable case that correspond to the sought bounds; these bounds automatically collapse to a precise number for the subclass of identifiable problems and queries.

The paper is organized as follows: after providing background material in Section~\ref{sec:back}, an algorithm to convert a Markovian causal model into a credal network, whose quantification is defined by the observational data, is provided in Section~\ref{sec:id}. The credal network mapping turns out to be extendible to quite a wider class of models that we call \emph{quasi-Markovian}; the scalability of the specification is discussed in the general case in Section \ref{sec:quasimarkov}. We regard this approach as the most systematic one presented so far in the literature and we discuss its advantages by a number of examples (Section \ref{sec:ex}). Finally the applicability to real-world cases is tested by numerical simulations showing that algorithms for approximate inference in credal networks allow to compute scalable and informative inferences even for large models (Section \ref{sec:exp}). We discuss future direction in Section~\ref{sec:conc}. 
\ifarxiv
Finally, the proofs of the theorems are in Appendix~\ref{sec:proofs}.
\fi

\section{Background Material}\label{sec:back}
\subsection{Bayesian and Credal Networks}
Let $X$ denote a variable taking values in a finite set $\Omega_X$. The elements of $\Omega_X$ are ordered and notation $x^{(k)}$ and $x$ is used, respectively, for the $k$-th and the generic element of $\Omega_X$. Denote as $P(X)$ a probability mass function (PMF) over $X$ and as $K(X)$ a \emph{credal} set (CS) over $X$, which is a set of PMFs over $X$. Given two variables $X$ and $Y$, a conditional probability table (CPT) $P(X|Y)$ is a collection of (conditional) PMFs indexed by the values of $Y$, i.e., $\{P(X|y)\}_{y\in\Omega_Y}$. If all PMFs in a CPT are \emph{degenerate}, i.e., there is a state receiving probability mass one and hence all the other ones receive zero, we say that also the CPT is degenerate. A credal CPT (CCPT) $K(X|Y)$ is similarly a collection of CSs over $X$ indexed by the values of $Y$. With a small abuse of terminology, we might call CPT (CCPT) also a single PMF (CS).

Consider a joint variable $\bm{X}:=(X_1,\ldots,X_n)$ and a directed acyclic graph $\mathcal{G}$ whose nodes are in one-to-one correspondence with the variables in $\bm{X}$ (whence we term a node in $\mathcal{G}$ and its corresponding variable interchangeably). Given $\mathcal{G}$, a Bayesian network (BN) is a collection of CPTs  $\{P(X_i|\mathrm{Pa}(X_i))\}_{i=1}^n$, where $\mathrm{Pa}(X_i)$ denotes the \emph{parents} of $X_i$, i.e., the direct predecessors of $X_i$ according to $\mathcal{G}$. A BN induces a joint PMF $P(\bm{X})$ that factorizes as follows: $P(\bm{x})=\prod_{i=1}^n P(x_i|\mathrm{pa}(X_i))$ for each $\bm{x}\in\Omega_{\bm{X}}$, where $(x_i,\mathrm{pa}(X_i))\sim \bm{x}$, i.e., $(x_i,\mathrm{pa}(X_i))$ are the values of $(X_i,\mathrm{Pa}(X_i))$ consistent with $\bm{x}$. A credal network (CN) is similarly intended as a collection of CCPTs. A CN defines a CS $K(\bm{X})$ whose elements are PMFs factorizing as those of a BN whose CPT values are taken from the corresponding CCPTs. Computing a conditional probability for a queried variable given an evidence with respect to the joint PMF of a BN, as well as the bounds of this probability with respect to the joint CS of a CN is NP-hard \citep{maua2014probabilistic}. Yet, polynomial algorithms computing approximate inferences for the general case (e.g., \citeauthor{approxlp}, \citeyear{approxlp}) or exact ones for classes of sub-models (e.g., \citeauthor{2u}, \citeyear{2u}) are available for CNs, not to mention the copious tools for BNs (e.g., \citeauthor{koller2009}, \citeyear{koller2009}).

\subsection{Structural Causal Models}\label{sec:scm}
The background concepts in this section are reviewed from the reference book of \citet{pearl2009causality}. Let us first define a \emph{structural equation} (SE) $f_X$ associated with variable $X$ and based on the input variable(s) $Y$ as a surjective function $f_X:\Omega_{Y} \to \Omega_X$ that determines the value of $X$ from that of $Y$. Consider two sets of variables $\bm{X}$ and $\bm{U}$ in one-to-one correspondence. We call \emph{endogenous} the variables in the first set and \emph{exogenous} the others. Say that, for each $X\in\bm{X}$, an SE $f_X$ associated with $X$ is provided. Assume that the exogenous variable $U$ corresponding to $X$ is an input variable of $f_X$, possibly together with other endogenous variables, but no other exogenous ones. We call a collection $\{f_X\}_{X\in\bm{X}}$ of SEs of this kind a \emph{structural causal model} $M$ (SCM) over $\bm{X}$. The \emph{causal diagram} $\mathcal{G}_M$ of an SCM $M$ is a directed graph over $(\bm{X},\bm{U})$ such that the parents $\mathrm{Pa}(X)$ of $X$ are the input variables of the SE $f_X$ for each $X\in\bm{X}$. We denote as $\underline{\mathrm{Pa}}(X)$ the endogenous parents of $X$ according to $\mathcal{G}_M$, i.e., $\underline{\mathrm{Pa}}(X):=\mathrm{Pa}(X) \setminus\{U\}$, for each $X\in\bm{X}$. 

Generally speaking $\mathcal{G}_M$ might include directed cycles. The SCM $M$ is \emph{semi-Markovian} if its causal diagram is acyclic. In a semi-Markovian SCM, a joint observation $\bm{U}=\bm{u}$ of the exogenous variables completely determines the state $\bm{X}=\bm{x}$ of the endogenous variables. This is achieved by obtaining the endogenous values from the SEs in a topological order for the variables in $\bm{X}$ according to $\mathcal{G}_M$. A \emph{probabilistic} SCM (PSCM) is a pair $(M,P)$ such that $M$ is a semi-Markovian SCM and $P$ a PMF over the exogenous variables $\bm{U}$. Exactly as we obtain $\bm{x}$ from $\bm{u}$ in SCMs, in PSCMs we obtain an endogenous PMF $P(\bm{X})$ from $P(\bm{U})$. PSCM $(M,P)$ is finally called \emph{Markovian} if its exogenous variables are jointly independent, i.e., $P(\bm{U})$ factorizes as $P(\bm{u})=\prod_{i=1}^n P(u_i)$ for each $\bm{u}\in\Omega_{\bm{U}}$ and $u_i\in\Omega_{U_i}$ with $u_i\sim\bm{u}$, for each $i=1,\ldots,n$.
Without loss of generality, we can intend non-Markovian PSCMs as based on SCMs whose exogenous variables might have multiple endogenous children, while keeping the joint independence for $P(\bm{U})$.\footnote{\cite{tian2002b} have shown that any SCM can be mapped to one whose $U$ variables are independent.} In the following we compactly describe $P$ in a PSCM by $\{P(U)\}_{U\in\bm{U}}$ in the place of $P(\bm{U})$. A Markovian PSCM is described here below, a non-Markovian one is in Example \ref{ex:nonmarkov}.

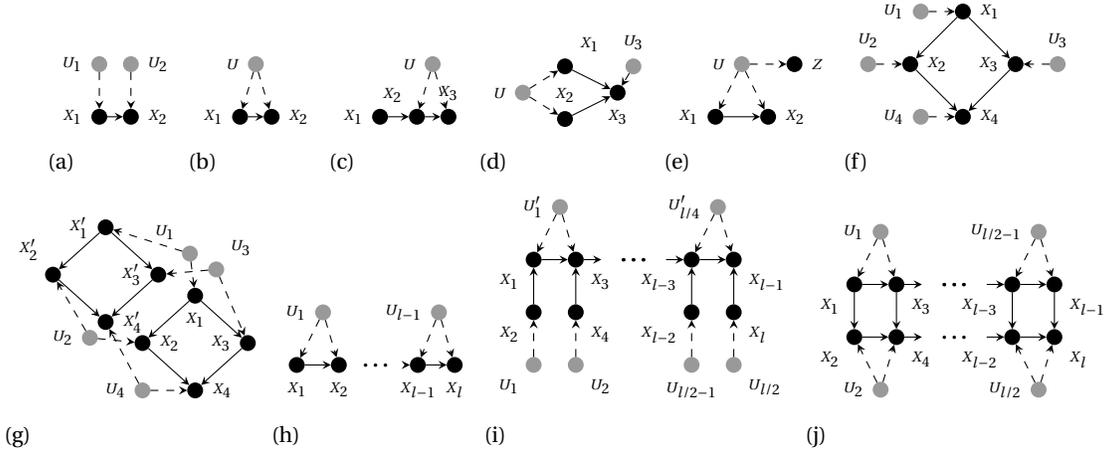
\begin{figure}[htp!]
\centering
\begin{subfigure}[]{}
\begin{tikzpicture}[scale=0.7]
\node[dot,label=left:{\tiny $X_1$}] (1)  at (0,0) {};
\node[dot,label=right:{\tiny $X_2$}] (2)  at (+0.6,0) {};
\node[dot2,label=left:{\tiny $U_1$}] (u1)  at (0,1) {};
\node[dot2,label=right:{\tiny $U_2$}] (u2)  at (+0.6,1) {};
\draw[a] (u1) -- (1);
\draw[a] (u2) -- (2);
\draw[a2] (1) -- (2);
\end{tikzpicture}
\end{subfigure}
\begin{subfigure}[]{}
\begin{tikzpicture}[scale=0.7]
\node[dot,label=left:{\tiny $X_1$}] (1)  at (0,0) {};
\node[dot,label=right:{\tiny $X_2$}] (2)  at (+0.6,0) {};
\node[dot2,label=left:{\tiny $U$}] (u)  at (0.3,1) {};
\draw[a] (u) -- (1);
\draw[a] (u) -- (2);
\draw[a2] (1) -- (2);
\end{tikzpicture}
\end{subfigure}
\begin{subfigure}[]{}
\begin{tikzpicture}[scale=0.7]
\node[dot,label=left:{\tiny $X_1$}] (1)  at (0.3,0) {};
\node[dot,label=above left:{\tiny $X_2$}] (2)  at (+1,0) {};
\node[dot,label=above:{\tiny $X_3$}] (3)  at (+1.6,0) {};
\node[dot2,label=left:{\tiny $U$}] (u)  at (1.3,1) {};
\draw[a] (u) -- (2);
\draw[a] (u) -- (3);
\draw[a2] (1) -- (2);
\draw[a2] (2) -- (3);
\end{tikzpicture}
\end{subfigure}
\begin{subfigure}[]{}
\begin{tikzpicture}[scale=0.7]
\node[dot,label=above right:{\tiny $X_1$}] (1)  at (1,1) {};
\node[dot,label=above:{\tiny $X_2$}] (2)  at (+1,0) {};
\node[dot,label=below:{\tiny $X_3$}] (3)  at (+2,0.5) {};
\node[dot2,label=left:{\tiny $U$}] (u)  at (0.2,.5) {};
\node[dot2,label=above:{\tiny $U_3$}] (u3)  at (2.3,1) {};
\draw[a] (u) -- (1);
\draw[a] (u) -- (2);
\draw[a] (u3) -- (3);
\draw[a2] (2) -- (3);
\draw[a2] (1) -- (3);
\end{tikzpicture}
\end{subfigure}
\begin{subfigure}[]{}
    \begin{tikzpicture}[scale=0.7]
    \node[dot,label=left:{\tiny $X_1$}] (1)  at (0,0) {};
    \node[dot,label=right:{\tiny $X_2$}] (2)  at (+1,0) {};
    \node[dot2,label=left:{\tiny $U$}] (u)  at (0.5,1) {};
    \node[dot,label=right:{\tiny $Z$}] (z)  at (+1.5,1) {};
    \draw[a] (u) -- (1);
    \draw[a] (u) -- (2);
    \draw[a] (u) -- (z);
    \draw[a2] (1) -- (2);
    \end{tikzpicture}
\end{subfigure}
\begin{subfigure}[]{}
\begin{tikzpicture}[scale=0.7]
\node[dot2,label=left:{\tiny $U_1$}] (u1)  at (-0.8,2) {};
\node[dot2,label=above:{\tiny $U_2$}] (u2)  at (-1.8,1) {};
\node[dot2,label=above:{\tiny $U_3$}] (u3)  at (1.8,1) {};
\node[dot2,label=left:{\tiny $U_4$}] (u4)  at (-0.8,0) {};
\node[dot,label=right:{\tiny $X_1$}] (1)  at (0,2) {};
\node[dot,label=right:{\tiny $X_2$}] (2)  at (-1,1) {};
\node[dot,label=left:{\tiny $X_3$}] (3)  at (1,1) {};
\node[dot,label=right:{\tiny $X_4$}] (4)  at (0,0) {};
\draw[a2] (1) -- (2);
\draw[a2] (1) -- (3);
\draw[a2] (3) -- (4);
\draw[a2] (2) -- (4);
\draw[a] (u1) -- (1);
\draw[a] (u2) -- (2);
\draw[a] (u3) -- (3);
\draw[a] (u4) -- (4);
\end{tikzpicture}
\end{subfigure}\\
\begin{subfigure}[]{}
\begin{tikzpicture}[scale=0.7]
\node[dot2,label=above left:{\tiny $U_1$}] (u1)  at (-0.1,2.6) {};
\node[dot2,label=left:{\tiny $U_2$}] (u2)  at (-2,1) {};
\node[dot2,label=above right:{\tiny $U_3$}] (u3)  at (0.4,2.3) {};
\node[dot2,label=left:{\tiny $U_4$}] (u4)  at (-1,0) {};
\node[dot,label=below:{\tiny $X_1$}] (1)  at (0,1.8) {};
\node[dot,label=right:{\tiny $X_2$}] (2)  at (-1,0.9) {};
\node[dot,label=left:{\tiny $X_3$}] (3)  at (1,0.9) {};
\node[dot,label=right:{\tiny $X_4$}] (4)  at (0,0) {};
\node[dot,label=left:{\tiny $X_1'$}] (11)  at (0-1.7,1.8+1.3) {};
\node[dot,label=above left:{\tiny $X_2'$}] (22)  at (-1-1.7,0.9+1.3) {};
\node[dot,label=left:{\tiny $X_3'$}] (33)  at (1-1.7,0.9+1.3) {};
\node[dot,label=right:{\tiny $X_4'$}] (44)  at (0-1.7,0+1.3) {};
\draw[a2] (1) -- (2);
\draw[a2] (1) -- (3);
\draw[a2] (3) -- (4);
\draw[a2] (2) -- (4);
\draw[a2] (11) -- (22);
\draw[a2] (11) -- (33);
\draw[a2] (33) -- (44);
\draw[a2] (22) -- (44);
\draw[a] (u1) -- (1);
\draw[a] (u1) -- (11);
\draw[a] (u2) -- (2);
\draw[a] (u2) -- (22);
\draw[a] (u4) -- (4);
\draw[a] (u4) -- (44);
\draw[a] (u3) -- (3);
\draw[a] (u3) -- (33);
\end{tikzpicture}
\end{subfigure}
    \begin{subfigure}[]{}
        \begin{tikzpicture}[scale=0.7]
        \node[dot,label=below:{\tiny $X_1$}] (1)  at (0,0) {};
        \node[dot,label=below:{\tiny $X_2$}] (2)  at (+0.8,0) {};
        \node[] (3)  at (+1.5,0) {$\;\;\ldots\;\;$};
        \node[dot,label=below:{\tiny $X_{l-1}$}] (Nm1)  at (+2.3,0) {};
        \node[dot,label=below:{\tiny $X_l$}] (N)  at (+3.,0) {};
        \node[dot2,label=left:{\tiny $U_1$}] (u12)  at (0.5,1) {};
        \node[dot2,label=left:{\tiny $U_{l-1}$}] (un)  at (2.7,1) {};
        \draw[a] (u12) -- (1);
        \draw[a] (u12) -- (2);
        \draw[a] (un) -- (Nm1);
        \draw[a] (un) -- (N);
        \draw[a2] (1) -- (2);
        \draw[a2] (3)--(Nm1);
        \draw[a2] (Nm1)--(N);
        \end{tikzpicture}
    \end{subfigure}
    \begin{subfigure}[]{}
        \begin{tikzpicture}[scale=0.7]
        \node[dot,label=below left:{\tiny $X_1$}] (1)  at (0,0) {};
        \node[dot,label=below right:{\tiny $X_3$}] (2)  at (+0.8,0) {};
        \node[] (3)  at (+1.9,0) {$\;\;\ldots\;\;$};
        \node[dot,label=below left:{\tiny $X_{l-3}$}] (Nm1)  at (+3.0,0) {};
        \node[dot,label=below right:{\tiny $X_{l-1}$}] (N)  at (+3.8,0) {};
        \node[dot2,label=left:{\tiny $U_1'$}] (u12)  at (0.5,1) {};
        \node[dot2,label=left:{\tiny $U_{l/4}'$}] (un)  at (3.5,1) {};
        \node[dot,label=below left:{\tiny $X_2$}] (y1)  at (0,-1) {};
        \node[dot,label=below right:{\tiny $X_4$}] (y2)  at (+0.8,-1) {};
        \node[] (3)  at (+1.9,0) {$\;\;\ldots\;\;$};
        \node[dot,label=below left:{\tiny $X_{l-2}$}] (yNm1)  at (+3.0,-1) {};
        \node[dot,label=below right:{\tiny $X_l$}] (yN)  at (+3.8,-1) {};
        \node[dot2,label=below left:{\tiny $U_1$}] (uy1)  at (0,-2) {};
        \node[dot2,label=below right:{\tiny $U_2$}] (uy2)  at (+0.8,-2) {};
        \node[dot2,label=below:{\tiny $U_{l/2-1}$}] (uyNm1)  at (+3.0,-2) {};
        \node[dot2,label=below right:{\tiny $U_{l/2}$}] (uyN)  at (+3.8,-2) {};	
        \draw[a] (u12) -- (1);
        \draw[a] (u12) -- (2);
        \draw[a] (un) -- (Nm1);
        \draw[a] (un) -- (N);
        \draw[a2] (1) -- (2);
        \draw[a2] (2)--(3);
        \draw[a2] (3)--(Nm1);
        \draw[a2] (Nm1)--(N);
        \draw[a2] (y1) -- (1);
        \draw[a2] (y2) -- (2);
        \draw[a2] (yNm1) -- (Nm1);
        \draw[a2] (yN) -- (N);
        \draw[a] (uy1) -- (y1);
        \draw[a] (uy2) -- (y2);
        \draw[a] (uyNm1) -- (yNm1);
        \draw[a] (uyN) -- (yN);
        \end{tikzpicture}
    \end{subfigure}
    \begin{subfigure}[]{}
        \begin{tikzpicture}[scale=0.7]
        \node[dot,label=below left:{\tiny $X_1$}] (1)  at (0,0) {};
        \node[dot,label=below right:{\tiny $X_3$}] (2)  at (+0.8,0) {};
        \node[] (3)  at (+1.9,0) {$\;\;\ldots\;\;$};
        \node[dot,label=below left:{\tiny $X_{l-3}$}] (Nm1)  at (+3.0,0) {};
        \node[dot,label=below right:{\tiny $X_{l-1}$}] (N)  at (+3.8,0) {};
        \node[dot2,label=left:{\tiny $U_1$}] (u12)  at (0.5,1) {};
        \node[dot2,label=left:{\tiny $U_{l/2-1}$}] (un)  at (3.5,1) {};
        \node[dot,label=below left:{\tiny $X_2$}] (y1)  at (0,-1) {};
        \node[dot,label=below right:{\tiny $X_4$}] (y2)  at (+0.8,-1) {};
        \node[] (3)  at (+1.9,0) {$\;\;\ldots\;\;$};
        \node[] (y3)  at (+1.9,-1) {$\;\;\ldots\;\;$};
        \node[dot,label=below left:{\tiny $X_{l-2}$}] (yNm1)  at (+3.0,-1) {};
        \node[dot,label=below right:{\tiny $X_l$}] (yN)  at (+3.8,-1) {};
        \node[dot2,label=left:{\tiny $U_{2}$}] (uy2)  at (+0.5,-2) {};
        \node[dot2,label=left:{\tiny $U_{l/2}$}] (uyN)  at (+3.5,-2) {};	
        \draw[a] (u12) -- (1);
        \draw[a] (u12) -- (2);
        \draw[a] (un) -- (Nm1);
        \draw[a] (un) -- (N);
        \draw[a2] (1) -- (2);
        \draw[a2] (2)--(3);
        \draw[a2] (3)--(Nm1);
        \draw[a2] (Nm1)--(N);
        \draw[a2] (y1) -- (y2);
        \draw[a2] (y2)--(y3);
        \draw[a2] (y3)--(yNm1);
        \draw[a2] (yNm1)--(yN);
        \draw[a2] (1) -- (y1);
        \draw[a2] (2) -- (y2);
        \draw[a2] (Nm1) -- (yNm1);
        \draw[a2] (N) -- (yN);
        \draw[a] (uy2) -- (y1);
        \draw[a] (uy2) -- (y2);
        \draw[a] (uyN) -- (yNm1);
        \draw[a] (uyN) -- (yN);
        \end{tikzpicture}
    \end{subfigure}
\caption{Ten causal diagrams.\label{fig:sm}}
\end{figure}

\begin{example}\label{ex:sm}
Consider two endogenous variables $(X_1,X_2)$ and their exogenous variables $(U_1,U_2)$. Let $\Omega_{X_i}:=\{\vx{i}k\}_{k=1}^2$  for $i=1,2$,  $\Omega_{U_1}:=\{\vu1k\}_{k=1}^3$ and
$\Omega_{U_2}:=\{\vu2k\}_{k=1}^5$. Define an SE for $X_1$ given $U_1$ such that $f_{X_1}(u_1^{(1)})=x_1^{(1)}$,  $f_{X_1}(u_1^{(2)})=x_1^{(2)}$, and
$f_{X_1}(u_1^{(3)})=x_1^{(2)}$. Similarly, for $X_2$ given $U_2$ and $X_1$, define an SE such that $f_{X_2}(u_2^{(k)},x_1^{(1)})=x_2^{(1)}$ for $k=3,4,5$ and $f_{X_2}(u_2^{(k)},x_1^{(2)})=x_2^{(1)}$ for $k=3,5$, while the other values of $k$ are giving $x_2^{(2)}$. The causal diagram $\mathcal{G}_M$ corresponding to this SCM is depicted in Figure \ref{fig:sm}.a. PSCM $(M,P)$ based on $M$ is obtained by also providing a PMF $P(U_1,U_2)$. Since this joint has to factorize into $P(U_1)P(U_2)$, as it follows from the graph $\mathcal{G}_M$, we know already that the SCM is Markovian. By finally taking both $P(U_1)$ and $P(U_2)$ uniform, we eventually get the PMF $P(X_1,X_2)$ such that $P(x_1^{(1)},x_2^{(1)})=\frac{1}{5}$, $P(x_1^{(1)},x_2^{(2)})=\frac{2}{15}$, and $P(x_1^{(2)},x_2^{(1)})=\frac{4}{15}$.
\end{example}

The following example shows that an SE $f_X$ in an SCM $M$ defines a degenerate CPT $P_M(X|\mathrm{Pa}(X))$.
\begin{example}\label{ex:cpts}
The SE $f_{X_1}$ of the SCM $M$ in Example~\ref{ex:sm} corresponds to the following CPT:
\begin{equation}
P_{M}(X_1|U_1)=
\begin{blockarray}{cccl}
\color{gray}{\vu11} &  \color{gray}{\vu12} & \color{gray}{\vu13} \\
\begin{block}{[ccc]l}
1 &0 &0 & {\color{gray}{\vx11} }\\
0 &1 &1 & \color{gray}{\vx12} \\
\end{block}
\end{blockarray}\,.
\end{equation}
Similarly, $P_{M}(X_2|U_1,x_1^{(1)})$, i.e., the restriction of the CPT of $X_2$ for $X_1=x_1^{(1)}$, is:
\begin{equation}\label{eq:cpt2}
\begin{blockarray}{cccccl}
\color{gray}{\vu21} &  \color{gray}{\vu22} & \color{gray}{\vu23} & \color{gray}{\vu24} & \color{gray}{\vu25}  \\
\begin{block}{[ccccc]l}
0&0&1&1&1&\color{gray}{\vx21}\\
1&1&0&0&0&\color{gray}{\vx21}\\
\end{block}
\end{blockarray}\,.
\end{equation}
\end{example}

From Example \ref{ex:cpts} and the joint independence of the exogenous variables, we have that a PSCM $(M,P)$ defines a joint PMF $P_{M}(\bm{X},\bm{U})$ such that $P_{M}(\bm{x},\bm{u})=\prod_{U\in\bm{U}} P(u) \cdot \prod_{X\in\bm{X}} P_{M}(x|\mathrm{pa}(X))$, for each $\bm{x}\in\Omega_{\bm{X}}$. This means that PSCMs are BNs whose endogenous degenerate CPTs are induced by the SEs of $M$, while the marginal PMFs over single exogenous variables are those in $P$.

Generally speaking SEs are non-injective and hence non-bijective maps. Yet, with a small abuse of notation, we denote as $f_X^{-1}$ the map $\Omega_X \to 2^{\Omega_{\mathrm{Pa}_X}}$ returning the set of values of the parents of $X$ corresponding to a particular value of $X$. E.g., in Example \ref{ex:sm}, $f_{X_1}^{-1}(x_1^{(2)})=\{u_1^{(2)},u_1^{(3)}\}$. For SEs of endogenous variables having both endogenous and exogenous parents, we perform such an inversion for a restriction of the SE obtained for given values of all the endogenous parents. E.g., the inverse of the SE for $X_2$ in Example \ref{ex:sm} given that $X_1=x_1^{(1)}$ is denoted as $f_{X_2|x_1^{(1)}}^{-1}$ and, following the table in Equation \eqref{eq:cpt2}, writes as $f^{-1}_{X_2|x_1^{(1)}}(x_2^{(1)})=\{u_2^{(3)},u_2^{(4)},u_2^{(5)}\}$. This notation for \emph{restricted inverse} maps of an SE will be used also in the general case.

\subsection{Interventions and Causal Effects}
The basic tool for causal analysis in an SCM $M$ is a mathematical operator called \emph{atomic intervention} and denoted as $\mathrm{do}(\cdot)$. Given $X \in \bm{X}$ and $x\in\Omega_X$, $\mathrm{do}(X=x)$ simulates a physical action on $M$ forcing $X$ to take a value $x\in\Omega_X$. Accordingly, the original SE $f_X$ should be replaced by a constant map $X=x$. Notation $M_x$ is used for such a modified SCM, whose causal diagram $\mathcal{G}_{M_x}$ can be obtained by removing from $\mathcal{G}_M$ the arcs entering $X$, and setting $X=x$ as an evidence. In a PSCM $(M,P)$, given $X,Y\in\bm{X}$ and $x\in\Omega_X$, we denote as $P_{M}(y|\mathrm{do}(x))$ the conditional probability of $Y=y$ in the post-intervention model $(M_x,P)$, i.e., $P_{M_x}(y|x)$, for each $y\in\Omega_Y$. The list $\{ P_{M}(y|\mathrm{do}(x))\}_{x\in\Omega_X}$ is called the \emph{causal effect} of $X$ on $Y=y$. As interventions commute, there are no ordering issues when coping with multiple interventions. If evidence is also available, i.e., some variables have been observed, it is customary to assume that observations took place after the interventions. 

\section{Coping with Unidentifiability}\label{sec:id}
Let us consider the problem of performing inference in a PSCM $(M,P)$. The exogenous variables $\bm{U}$ are typically assumed to be latent, i.e., directly unobservable, while the endogenous variables $\bm{X}$ are manifest. Assume for a moment that their joint PMF $\tilde{P}(\bm{X})$ is known. Following the discussion in Section \ref{sec:scm}, $\tilde{P}(\bm{X})$ is just the image, through the SEs of $M$, of the latent PMFs $\{P(U)\}_{U\in\bm{U}}$. 

The focus of this paper in primarily on the problem of learning the (latent) latter from the (manifest) former: that is, given $\tilde{P}(\bm{X})$ and an SCM $M$ over $\bm{X}$, to find out $\mathcal{K}_{M,\tilde{P}}$, namely, the collection of CSs $\{K(U)\}_{U\in\bm{U}}$ inducing PSCMs based on $M$ such that:
\begin{equation}\label{eq:ident}
\sum_{u\in\Omega_U, U\in\bm{U}} \left[ \prod_{U\in\bm{U}} P(u) \cdot \prod_{X\in\bm{X}} P_M(x|u,\underline{\mathrm{pa}}(X)) \right] = \tilde{P}(\bm{x})\,,
\end{equation}
for each $\bm{x}\in\Omega_{\bm{X}}$, $(x,\underline{\mathrm{pa}}(X))\sim \bm{x}$, with $P(U)\in K(U)$ for each $U\in\bm{U}$. Once $\{K(U)\}_{U\in\bm{U}}$ have been defined, our next aim is to use them to make causal inference on a generic real-valued quantity $Q$ that can be obtained from a PSCM $(M,P)$. In particular, we aim at computing the \emph{causal bounds} of $Q$: $[\min_{P\in \mathcal{K}_{M,\tilde{P}}} Q(M,P),\max_{P\in \mathcal{K}_{M,\tilde{P}}} Q(M,P)]$. If this interval reduces to a point, we say that $Q$ is \emph{identifiable} in $\mathcal{K}_{M,\tilde{P}}$; otherwise it is called \emph{unidentifiable}.

Note that in practice we will most likely not have the exact joint $\tilde{P}(\bm{X})$, but rather a sample $\mathcal{D}$ from such a PMF; as a consequence $\tilde{P}(\bm{X})$ will be an approximation to the actual PFM and for this reason we shall call it `empirical' in the following. We shall also assume it to be strictly positive. 

\subsection{Causal Bounds by Credal Networks}
The collection of CSs $\mathcal{K}_{M,\tilde{P}}$ defined in the previous section provides a parametrization of PSCMs compatible with Equation \eqref{eq:ident}. We call $\mathcal{K}_{M,\tilde{P}}$ the \emph{identification} through $M$ of $\tilde{P}$. It is worth noticing that, as well as a PSCM corresponds to a BN, the PSCMs induced by $\mathcal{K}_{M,\tilde{P}}$ correspond to a CN, this allowing to address causal bound computation by standard inference algorithms for CNs. An example is here below.

\begin{example}\label{ex:single}
Consider an SCM $M$ made of a single endogenous variable $X$ and a single exogenous variable $U$. Given an empirical PMF $\tilde{P}(X)$, consider a CN over $\mathcal{G}_M$ such that the CCPT of $X$ is the  (degenerate) CPT $P_M(X|U)$ induced by $f_X$, and the CCPT of $U$ is the CS $K(U)$ induced by the linear constraints on $P(U)$:
\begin{equation}\label{eq:constraint}
\sum_{u \in f_X^{-1}(x)} P(u) = \tilde{P}(x)\,,
\end{equation}
for each $x\in\Omega_X$.
 $K(U)$ coincides with the identification $\mathcal{K}_{M,\tilde{P}}$, i.e., every quantification of $P(U)$ consistent with Equation \eqref{eq:constraint} solves Equation \eqref{eq:ident} and vice versa. To check that, let us compute the marginal probabilities over $X$ in a PSCM $(M,P)$ with $P\in \mathcal{K}_{M,\tilde{P}}$, i.e.,
\begin{equation}\label{eq:constraint2}
P(x) = \sum_{u \in \Omega_U} P_M(x|u) \cdot P(u) =
\sum_{u \in \Omega_U} 
\llbracket f_X(u)=x \rrbracket \cdot P(u)\,,
\end{equation}
where the first step is by total probability theorem, the second follows from the fact that $P_M(X|U)$ is a degenerate CPT, and $\llbracket \cdot \rrbracket$ is an Iverson bracket giving one if its argument is true and zero otherwise. As the rightmost-hand side of Equation \eqref{eq:constraint2} coincides with the left-hand side of Equation \eqref{eq:constraint}, we have the one-to-one mapping between the elements of $\mathcal{K}_{M,\tilde{P}}$ and those of $K(U)$. For a numerical example, consider a ternary $U$ and a binary $X$ whose SE coincides with $f_{X_1}$ in Example \ref{ex:sm}. For $\tilde{P}(x^{(1)})=\frac{1}{3}$, Equation \eqref{eq:constraint} gives $P(u^{(1)})=\frac{1}{3}$ and $P(u^{(2)}) + P(u^{(3)})=\frac{2}{3}$, i.e., any $P(U):=[\frac{1}{3}, p, \frac{2}{3}-p]$ with $p \in [0,\frac{2}{3}]$, and no other one, is consistent with $\tilde{P}$.
\end{example}

\subsection{Markovian Case}\label{sec:markov}
The procedure in Example \ref{ex:single} can be extended to any SCM under the Markovian assumption. For an endogenous variable $X$ whose endogenous $U$ is the unique parent, the procedure is exactly the same, and the constraints on the marginal probabilities of $U$ are like those in Equation \eqref{eq:constraint}. If $X$ has other, endogenous, parents besides $U$, more analysis is required to write the analogous of Equation \eqref{eq:constraint}. In this case, the SE of $X$ has form  $x=f_X(u,\underline{\mathrm{pa}}(X))$. Each $\underline{\mathrm{pa}}(X)\in\Omega_{\underline{\mathrm{Pa}}(X)}$ induces a restricted inverse of $f_X$ and hence, for each $x\in\Omega_X$, the analogous of Equation \eqref{eq:constraint} becomes:
\begin{equation}\label{eq:constraint_parents}
\sum_{u \in f^{-1}_{X|\underline{\mathrm{pa}}(X)}(x)} P(u)=\tilde{P}(x|\underline{\mathrm{pa}}(X))\,,
\end{equation}
where the conditional probabilities on the right-hand side are obtained from the empirical PMF $\tilde{P}(\bm{X})$. These are constraints on the elements of PMF $P(U)$ that can be applied separately, for each $U\in\bm{U}$, because of the Markovian assumption.
The procedure is detailed by Algorithm \ref{alg:simple}.

\begin{algorithm}[htp!]
\caption{Given an SCM $M$ and a PMF $\tilde{P}(\bm{X})$, return CSs $\{K(U)\}_{U\in\bm{U}}$\label{alg:simple}}
\begin{algorithmic}[1]
\FOR {$X\in\bm{X}$}
\STATE $U \leftarrow \mathrm{Pa}(X) \cap \bm{U}$
{\hspace*{\fill}// $U$ as the unique exogenous parent of $X$}
\STATE $\underline{\mathrm{Pa}}(X) \leftarrow \mathrm{Pa}(X) \setminus \{ U \}$
{\hspace*{\fill}// Endogenous parents of $X$}
\IF{$\underline{\mathrm{Pa}}(X) = \emptyset$}
\STATE $K(U) \leftarrow \{P'(U)\,:\, 
\sum_{u \in f^{-1}_{X}} P'(u)=\tilde{P}(x)\,,
\forall x\in\Omega_X\}$
{\hspace*{\fill}// Eq.~ \eqref{eq:constraint}}
\ELSE
\STATE $K(U) \leftarrow \{P'(U)\,:\, 
\sum_{u \in f^{-1}_{X|\underline{\mathrm{pa}}(X)}(x)} P'(u)=\tilde{P}(x|\underline{\mathrm{pa}}(X))\,,
\forall x\in\Omega_X , \forall \underline{\mathrm{pa}}(X)\in\Omega_{\underline{\mathrm{Pa}}_X}\}$
{\hspace*{\fill}// Eq.~ \eqref{eq:constraint_parents}}
\ENDIF
\ENDFOR
\end{algorithmic}
\end{algorithm}

Overall, this allows to compute the identification through $M$ of $\tilde{P}$ in the Markovian case.

\begin{theorem}\label{th:markov}
In the Markovian case, the output $\{K(U)\}_{U\in\bm{U}}$ of Algorithm \ref{alg:simple} coincides with $\mathcal{K}_{M,\tilde{P}}$.
\end{theorem}


\begin{example}\label{ex:markov}
Consider the Markovian SCM $M$ in Example \ref{ex:sm} whose PMF $P(X_1,X_2)$ is used as empirical PMF $\tilde{P}(X_1,X_2)$. We obtain $\tilde{P}(\vx11)=\frac{1}{3}$,  $\tilde{P}(\vx21|\vx11)=\frac{3}{5}$, and $\tilde{P}(\vx21|\vx12)=\frac{2}{5}$. In this setup Algorithm \ref{alg:simple} returns a CS $K(U_1)$ equal to $K(U)$ in Example~\ref{ex:single}, while Equation \eqref{eq:constraint_parents} for $U_2$ gives:
\begin{equation}
\begin{array}{lll}
P(\vu23) + P(\vu24) + P(\vu25) &= P(\vx21|\vx11)=0.6\,,\\
 P(\vu23) + P(\vu25) &= P(\vx21|\vx12)=0.4\,,\\
P(\vu21) + P(\vu22)  &= P(\vx22|\vx11)=0.4\,,\\
P(\vu21) + P(\vu22) + P(\vu24) &= P(\vx22|\vx12)=0.6\,.\\
\end{array}
\end{equation}

This defines a CS $K(U_2)$ equivalent to any convex combination of the PMFs:
\begin{equation}
P_1(U_2) =
\begin{blockarray}{[c]}
0.0 \\  0.4 \\  0.4 \\  0.2 \\  0.0\\
\end{blockarray},
P_2(U_2)=
\begin{blockarray}{[c]}
0.4 \\  0.0 \\  0.4 \\  0.2 \\  0.0\\
\end{blockarray},
P_3(U_2)=
\begin{blockarray}{[c]}
0.0 \\  0.4 \\  0.0 \\  0.2 \\  0.4\\
\end{blockarray}\,,
P_4(U_2)=
\begin{blockarray}{[c]}
0.4 \\  0.0 \\  0.0 \\  0.2 \\  0.4\\
\end{blockarray}\,.
\end{equation}
As expected the uniform PMF over $U_1$ in Example \ref{ex:sm} is included in $K(U_1)$ and the same happens for $U_2$ and $K(U_2)$.
\end{example}

\section{Beyond Markovianity}\label{sec:quasimarkov}
We extend the tools of the previous section to non-Markovian models starting from an example.

\begin{example}\label{ex:nonmarkov}
Consider an SCM $M$ over two binary endogenous variables $X_1$ and $X_2$ whose common exogenous parent $U$ has five states. 
The SE for $X_1$ is such that $f_{X_1}(\vu{{}}k) = \vx11$ for $k=1,4,5$ and $\vx12$ otherwise. 
For $X_2$, we have instead $f_{X_2}(\vu{{}}k , \vx11) = \vx21$ for $k=1,3$, $f_{X_2}(\vu{{}}k , \vx11) = \vx22$ for $k=2,4,5$, 
$f_{X_2}(\vu{{}}k , \vx12) = \vx21$ for $k=3$, and 
$f_{X_2}(\vu{{}}k , \vx12) = \vx22$ for $k=1,2,4,5$. The causal diagram of $M$ is the one in Figure \ref{fig:sm}.b. A (non-Markovian) PSCM based on $M$ would be obtained by any specification of PMF $P(U)$.
\end{example}


In the non-Markovian case, the common exogenous parents of two or more endogenous variables are called \emph{confounders}. Confounders express non-Markovianity also at the SCM level, being input variables common to two or more SEs. In these cases, the surjectivity we assume for single SEs is extended to the joint SE involving all the SEs with the same confounder in input.

\begin{example}\label{ex:nonmarkov2}
Let $\tilde{P}(X_1,X_2)$ denote the empirical PMF of the non-Markovian PSCM $(M,P)$ in Example \ref{ex:nonmarkov}, whose causal diagram is in Figure \ref{fig:sm}.b.  In this case, Equation \eqref{eq:ident} rewrites as:
\begin{equation}\label{eq:constraint33}
\sum_{u \in \Omega_U} P(u) P_M(x_1|u) P_M(x_2|u) = \tilde{P}(x_1,x_2)\,,
\end{equation}
to be considered for each $x_1\in\Omega_{X_1}$ and $x_2\in\Omega_{X_2}$. In the sum on the left-hand side, the values of $u$ that are not simultaneously consistent, through SEs $f_{X_1}$ and $f_{X_2}$, with both $x_1$ and $x_2$, are zero. We therefore rewrite Equation \eqref{eq:constraint33} as:
\begin{equation}\label{eq:constraint3}
\sum_{u \in f_{X_1}^{-1}(x_1) \cap f_{X_2}^{-1}(x_2)}P(u)=\tilde{P}(x_1,x_2)\,.
\end{equation}

As a numerical example consider the empirical PMF $\tilde{P}(X_1,X_2)$ in Example~\ref{ex:sm}. The corresponding constraints for $U$ are: $P(\vu{{}}1) = \frac{1}{5}$, $P(\vu{{}}2) = \frac{2}{5}$ and  $P(\vu{{}}3) = \frac{4}{15}$, and $P(\vu{{}}4) + P(\vu{{}}5) = \frac{2}{15}$. The elements of the corresponding CS $K(U)$ are therefore
$P(U) = [\frac{1}{5},\frac{2}{5},\frac{2}{15},t,\frac{1}{15}-t]$ with $t\in[0,\frac{1}{15}]$.
\end{example}

A procedure analogous to that in Example \ref{ex:nonmarkov2} can be used for the identification of $\tilde{P}$ for non-Markovian SCMs. As in the previous section, the key point is that the constraints on the marginal PMF of an exogenous variable $U\in\bm{U}$ imposed by the consistency with the empirical PMF can be specified separately from those of the other exogenous variables. We call \emph{quasi-Markovian} a PSCM $(M,P)$ such that each $X\in\bm{X}$ has only a single $U\in\bm{U}$ as parent.\footnote{In principle any semi-Markovian model can be turned into quasi-Markovian, e.g., by clustering all $U$ variables into a single one; yet this neglects the exponential blowup in the computation that follows as a consequence.} In other words, in a quasi-Markovian model, if $X$ is a child of $U$, we have $\mathrm{Pa}(X)=(U,\underline{\mathrm{Pa}}(X))$ where, despite the possible non-Markovianity, $\underline{\mathrm{Pa}}(X)\subset\bm{X}$. This, together with the joint independence of the PMFs $\{P(U)\}_{U\in\bm{U}}$, induces in quasi-Markovian PSCMs the factorization:
\begin{equation}\label{eq:qmfact}
P(\bm{x},\bm{u}) = \prod_{U \in\bm{U}} \left[ P(u) \prod_{X \in \bm{X}: U \in \mathrm{Pa}(X)} P(x|\underline{\mathrm{pa}}(X),u)
\right]\,,
\end{equation}
for each $\bm{x}\in\Omega_{\bm{X}}$ and $\bm{u}\in\Omega_{\bm{U}}$. It is easy to see that, because of the quasi-Markovianity, in the product over $\bm{U}$ in the right-hand side of Equation \eqref{eq:qmfact}, the states of each $U\in\bm{U}$ appear only in the corresponding factor.
This allows to define a procedure analogous to Algorithm \ref{alg:simple} to derive the CSs $\{K(U)\}_{U\in\bm{U}}$ and hence obtain a CN from a quasi-Markovian model.\footnote{Algorithm~\ref{alg:quasi} and Theorem~\ref{th:quasi} are tightly related to the notion of \emph{confounded component} by \citet{tian2002general}.}

\begin{algorithm}[htp!]
\caption{Given an SCM $M$ and a PMF $\tilde{P}(\bm{X})$, return CSs $\{K(U)\}_{U\in\bm{U}}$\label{alg:quasi}}
\begin{algorithmic}[1]
\FOR{$U\in\bm{U}$}
\STATE $\{X_U^k\}_{k=1}^{n_U} \leftarrow \mathrm{Sort}[X \in \bm{X} : U \in \mathrm{Pa}(X)]$
{\hspace*{\fill}// Children of $U$ in topological order}
\STATE $\gamma \leftarrow \emptyset$
\FOR{$(x_U^1,\ldots,x_U^{n_U}) \in \times_{k=1}^{n_U} \Omega_{\bm{X}_U^k}$}
\FOR{$(\underline{\mathrm{pa}}(X_U^1),\ldots,\underline{\mathrm{pa}}(X_U^{n_U})) \in \times_{k=1}^{n_U} \Omega_{\underline{\mathrm{Pa}}(X_U^k)}$}
\STATE{$\Omega_U' \leftarrow  \bigcap_{k=1}^{n_U} f^{-1}_{X_U^k|\underline{\mathrm{pa}}({X_U^k})}(x_U^k)$}
\STATE{ $\gamma \leftarrow \gamma \cup \left\{ \sum_{u\in\Omega_U'} P(u) = \prod_{k=1}^{n_U} \tilde{P}(x_U^k|x_U^1,\ldots,x_U^{k-1}, \underline{\mathrm{pa}}({X_U^1})),\ldots,\underline{\mathrm{pa}}({X_U^k})) \right\}$}
\ENDFOR
\ENDFOR
\STATE{ $K(U) \leftarrow \{ P(U): \gamma \}$}
{\hspace*{\fill}// CS by linear constraints on $P(U)$}
\ENDFOR
\end{algorithmic}
\end{algorithm}

Algorithm \ref{alg:quasi} computes the CN representation of a quasi-Markovian SCM. It is a simple exercise to check that the algorithm coincides with Algorithm \ref{alg:simple} in the case of Markovian inputs (as $n_U=1$ for each $U\in\bm{U}$). In fact, Algorithm \ref{alg:quasi} can be regarded as a (partial) extension to the non-Markovian case of Algorithm \ref{alg:simple} as shown by the following result, analogous to Theorem \ref{th:markov}.

\begin{theorem}\label{th:quasi}
In quasi-Markovian cases, the output $\{K(U)\}_{U\in\bm{U}}$ of Algorithm \ref{alg:quasi} coincides with $\mathcal{K}_{M,\tilde{P}}$.
\end{theorem}

Regarding complexity, unlike Algorithm \ref{alg:simple}, where the size of the CSs constraints of the CN are the same as the SEs in the SCM, the bottleneck of Algorithm \ref{alg:quasi} are the two nested loops (lines 4--9). This roughly corresponds to a complexity exponential with respect to the maximum number of children of the exogenous variables. Setting a bound to this number (e.g., $n_U \leq 2$, meaning that confounders only act on pairs of endogenous variables), would make the procedure polynomial for quasi-Markovian models.

\section{Causal Analysis by Credal Networks}\label{sec:ex}
In the previous sections we proved that CNs exactly represent the uncertainty about the exogenous variables of an SCM induced by an empirical endogenous PMF $\tilde{P}(\bm{X})$. Such an abstract result allows us in practice to bound causal inference. Given a (quasi-)Markovian SCM $M$ and an empirical PMF $\tilde{P}(\bm{X})$, we first obtain its CN representation by Algorithms \ref{alg:simple} or \ref{alg:quasi}. Post-interventional queries in $M$ are then addressed via the CN, since it preserves the separation properties of $M$; the degenerate (C)CPTs of the SEs involved in the intervention are replaced by constants (\emph{surgery}), and the queries are then obtained by standard CN  algorithms. Let us start from the identifiable case, where post-intervention queries are reduced to pre-intervention ones.

\begin{example}[Backdoor Identification \citep{pearl2009causality}]
In the non-Markovian SCM in Figure \ref{fig:sm}.d, consider the query $P(x_3|\mathrm{do}(x_1))$. The intervention on $X_1$ requires the removal of the arc from $U$ towards $X_1$, and hence:
\begin{equation}\label{eq:docalculus}
P(x_3|\mathrm{do}(x_1)):=\sum_{x_2,u,u_3} P(x_3|x_1,x_2,u_3) P(u_3) P(x_2|u) P(u) =\sum_{x_2} P(x_3|x_1,x_2) P(x_2)\,.
\end{equation}
In the CN obtained for such quasi-Markovian model, the task becomes the computation of the lower and upper bounds of the second term in Equation \eqref{eq:docalculus} when $P(U)\in K(U)$ and $P(U_2)\in K(U_2)$, where the CSs are obtained by Algorithm \ref{alg:quasi}. As the CN representation is equivalent, the CN inference gives equal upper and lower bounds, which expresses query identifiability.
\end{example}

For identifiable tasks, our approach offers then a numeric alternative to analytical approaches such as \emph{do calculus} \citep{pearl1995causal}. Advantages become more evident in non-identifiable cases:

\begin{example}[Unidentifiable Clinical Trials \citep{balke1994counterfactual}]\label{ex:trial}
In the SCM in Figure \ref{fig:sm}.c, the endogenous variables are binary. As $f_{X_1}$ implements $U_1 = X_1$, we do not explicitly show $U_1$. Regarding $U$, $|\Omega_U|=16$ and the SEs $f_{X_2}$ and $f_{X_3}$ are as in the original example (not reported here for lack of space). The empirical PMF is such that $\tilde{P}(x_1^{(2)})=0.1$, $\tilde{P}(X_2,X_3|x_1^{(1)})=[0.32,0.32,0.04,0.32]$, and
$\tilde{P}(X_2,X_3|x_1^{(2)})=[0.02,0.17,0.67,0.14]$. Algorithm \ref{alg:quasi} gives the constraints for $K(U)$ and the CN computes the causal bounds
 $P(x_3^{(2)}|\mathrm{do}(x_2^{(1)}))\in[0.45,0.46]$ and 
 $P(x_3^{(2)}|\mathrm{do}(x_2^{(2)}))\in[0.67,0.68]$. The corresponding difference between the two effects is the interval $[-0.23,-0.21]$, which is narrower than $[-0.23,-0.15]$, the interval obtained by \citet{balke1994counterfactual} with linear programming.
\end{example}

Our procedure should be therefore regarded as a numerical counterpart of the symbolic approach of \citet{balke1994counterfactual}, 
recently extended to more general cases by \citet{sachs2020}. Apart from their restriction to binary variables only, the linear programming reduction in those approaches relaxes some of the constraints on the exogenous PMFs, thus eventually yielding an approximate characterization of the bounds. In practice, as seen in the above example, the bounds provided by these programs will be an outer approximation of the exact ones we provide.

Our CN equivalence results can be also applied to counterfactuals, namely queries that represent more than one `world' at the same time, the actual one and other, alternative worlds. \cite{balke1994cf} showed how to compute these queries in PSCMs by \emph{twin nets} (more generally called `counterfactual graphs'). In these models, each endogenous variable has a replica, sharing the same exogenous parents and the SEs. The CSs returned by Algorithms \ref{alg:simple} or \ref{alg:quasi} can be also used in twin nets, thus allowing us to bound such queries on the basis of empirical data.

\begin{example}[What-If At the Party \citep{balke1994cf}]\label{ex:party}
In the Markovian SCM of Figure \ref{fig:sm}.f, the binary endogenous variables $(X_1,X_2,X_3,X_4)$ have exogenous counterparts with cardinalities: $|\Omega_{U_1}|=2$, $|\Omega_{U_2}|=|\Omega_{U_3}|=4$, and $|\Omega_{U_4}|=3$. We refer to the original paper for the specification of the SEs and $\{P(U_k)\}_{k=1}^4$. For the corresponding PSCM, in that paper, the twin net is built as in Figure~\ref{fig:sm}.g and used to compute the counterfactual $P({x_4}^{(2)}_{x_3^{(2)}}|x_3^{(1)})=0.79$, modelling the hypothetical effect on $X_4$ of an intervention on $X_3$ forcing a state different from the observed one. We address the same task without exploiting $\{P(U_k)\}_{k=1}^4$, using only the empirical PMF. Algorithm \ref{alg:simple} can compute CSs $\{K(U_k)\}_{k=1}^4$ in the original model, and use them to make the twin net as a CN. In spite of the looser information, the above counterfactual in the twin CN takes exactly the same sharp value.
\end{example}

\begin{myremark}[\bf Counterfactuals without structural equations]\label{rem:noinfo}The same can be done even if no explicit information about the SEs is available (this follows straightforwardly from ideas by, e.g., \citealt[Section~3.1]{balke1994counterfactual}). Consider a SE $x=f_X(x',u)$. The number of degenerate CPTs $P(X|X')$ modelling a deterministic relation between $X'$ and $X$ is $k:=|\Omega_X|^{|\Omega_{X'}|}$. Accordingly, we set $\Omega_U:=k$ and let $P(X|X',u^{(k)})$ correspond to the $k$-th degenerate CPT $P(X|X')$. Modelling ignorance about SE $f_{X_4}$ in Example \ref{ex:party} requires for instance $|\Omega_{U_4}|:=|\Omega_{X_4}|^{|\Omega_{X_2}|\cdot|\Omega_{X_3}|}=16$. By letting $U_4$ enumerate all the functional relations between $(X_2,X_3)$ and $X_4$ and keeping the same empirical PMF, we still obtain the same counterfactual value. To make the query unidentifiable, we add a $0.01$ cut-off to the degenerate values of $\tilde{P}$; this induces the bounds $P({x_4}^{(2)}_{x_3^{(2)}}|x_3^{(1)}) \in [0.75,0.85]$.
\end{myremark}

Many other tools for causal modelling and analysis can easily be embedded in our CN formalism. This is for instance the case of \emph{measurement bias} considered by \cite{pearl2012measurement}. A likelihood $\{P(z|U)\}_{z\in\Omega_Z}$ modelling a noisy observation of an exogenous variable $U$ might be available under the assumption of conditional independence between $Z$ and the other variables given $U$. This can be modelled by setting $Z$ as a binary child of $U$ (e.g., see Figure \ref{fig:sm}.e) in the CN, and using the likelihood for the specification of its CPT exactly as in Pearl's \emph{virtual evidence} method.

Expert judgements about $U$ (e.g., a comparative statement such as $P(u^{(i)})>P(u^{(j)})$) can be trivially embedded in our model as additional constraints in line 10 of Algorithm \ref{alg:quasi}.  Finally note also that, exactly as we perform standard surgery in the CN by exploiting the fact that the CN maintains the same separation properties of the original model, also other state-of-the art techniques such as the \emph{non-atomic interventions} proposed by \citet{correa2020calculus} could be supported, thus providing a numerical alternative to their $\sigma$-calculus.

\section{Numerical Tests}\label{sec:exp}
For validation, we use quasi-Markovian SCMs of increasing size and different topologies (of the endogenous restriction of the causal diagram): \emph{trees} (Figure \ref{fig:sm}.h), \emph{polytrees} (Figure \ref{fig:sm}.i), \emph{multiply connected} (Figure \ref{fig:sm}.j). We call  \emph{length} of the model $l:=|\bm{X}|$ the number of endogenous variables in $M$. We assume \emph{stationarity}, i.e., the same SEs are associated with the same variables with different indexes. A random $P(\bm{U})$ is sampled, and the corresponding empirical PMF $\tilde{P}(\bm{X})$ obtained. The latter is the input of Algorithm \ref{alg:quasi}. Endogenous and exogenous variables have resp. two and six states, while SEs are randomly generated. As unidentifiable queries, we consider
 $P(X_{l/2}|\mathrm{do}(x^{(0)}_1),x^{(0)}_l)$ for trees, $P(X_{l/4}|\mathrm{do}(x^{(0)}_1), x^{(0)}_{l-1})$ for polytrees, and $P(X_{l/4}|\mathrm{do}(x^{(0)}_1), x^{(0)}_l)$ for multiply connected models. Descriptors reported here are averages of $100$ iterations.
 
\begin{figure}[htp!]
\centering
\begin{tikzpicture}[yscale=.35,xscale=.45]
\begin{groupplot}[group style={group size=3 by 1, vertical sep=5cm, horizontal sep=0.2cm}]
\nextgroupplot[
tick align=outside,
tick pos=left,
title={Trees},
x grid style={white!69.0196078431373!black},
xlabel={Length $l$},
xmajorgrids,
xmin=1.95, xmax=21.05,
xtick style={color=black},
y grid style={white!69.0196078431373!black},
ylabel={Execution time (s)},
ymajorgrids,
ymin=-5.743513777, ymax=120.661044437,
ytick style={color=black}
]
\addplot [ultra thick,black,dotted]
table {%
3 0.34175796
4 0.4661655
5 0.74143504
6 0.8607
7 1.36857902
8 1.39887822
9 2.1185295
10 2.30123888
11 2.91466794
12 3.0599051
13 3.92544538
14 3.95807788
15 4.86445886
16 5.09564466
17 5.75508222
18 6.1757121
19 6.87571724
20 6.97033168
21 7.8040553
};
\addplot [ultra thick,black]
table {%
3 0.07276482
4 0.06845506
5 0.114651755555556
6 0.107642844444444
7 2.203501675
8 0.148208425
9 0.234517066666667
10 0.2576461
11 15.4956344666667
12 0.7151094
13 8.09172163333333
14 4.47642063333333
15 19.1777441333333
16 20.4822364
17 82.0155528666667
18 114.9153827
};

\nextgroupplot[
scaled y ticks=manual:{}{\pgfmathparse{#1}},
tick align=outside,
tick pos=left,
title={Polytree},
x grid style={white!69.0196078431373!black},
xlabel={Length $l$},
xmajorgrids,
xmin=1.95, xmax=21.05,
xtick style={color=black},
y grid style={white!69.0196078431373!black},
ylabel={ },
ymajorgrids,
ymin=-5.743513777, ymax=120.661044437,
ytick style={color=black},
yticklabels={},
]
\addplot [ultra thick,black,dotted]
table {%
6 1.45962308
8 2.56224282
10 3.73061524
12 4.74696154
14 6.39394936
16 7.9910649
18 9.42139992
20 10.6053752
};
\addplot [ultra thick,black]
table {%
6 0.203544177777778
8 0.305260822222222
10 2.29887262222222
12 11.4464603555556
};

\nextgroupplot[
legend cell align={left},
legend pos=north west,
legend style={fill opacity=0.8, draw opacity=1, text opacity=1, draw=white!80!black},
scaled y ticks=manual:{}{\pgfmathparse{#1}},
tick align=outside,
tick pos=left,
title={Multiply connected},
x grid style={white!69.0196078431373!black},
xlabel={Length $l$},
xmajorgrids,
xmin=1.95, xmax=21.05,
xtick style={color=black},
y grid style={white!69.0196078431373!black},
ylabel={ },
ymajorgrids,
ymin=-5.743513777, ymax=120.661044437,
ytick style={color=black},
yticklabels={},
]
\addplot [ultra thick,dotted]
table {%
6 1.9700621
8 2.34051874
10 4.7520846
12 4.90502024
14 7.84319146
16 8.4303279
18 12.8570035
20 13.15327414

};
\addlegendentry{ApproxLP}
\addplot [black,ultra thick]
table {%
6 0.148006
8 0.16463044
10 0.7672054
12 1.3035927
14 24.4201753
16 37.4387636
};
\addlegendentry{CVE}
\end{groupplot}
\end{tikzpicture}
\begin{tikzpicture}[xscale=0.45,yscale=0.35]
\begin{axis}[
legend cell align={left},
legend style={fill opacity=0.8, draw opacity=1, text opacity=1, draw=white!80!black},
tick align=outside,
tick pos=left,
title={ },
x grid style={white!69.0196078431373!black},
xlabel={Length $l$},
xmajorgrids,
xmin=4, xmax=16,
xtick style={color=black},
            yticklabel pos=right,
y grid style={white!69.0196078431373!black},
ylabel={$\max{P}-\min {P}$},
ymajorgrids,
ylabel near ticks,
ymin=0.0, ymax=0.1,
ytick={0, 0.05, 0.1},
yticklabels={$0.00$, $0.05$, $0.10$},
ytick style={color=black}]
\addplot [mark=triangle,black]
table {%
     6 0.016428715834721
    8 0.0160996738456202
    10 0.01
    12 0.0160545028126254
    15 0.01
};
\addlegendentry{Muliply connected}
\addplot [thick, black, dashed, mark=*]
table {%
	6 0.000514007273336123
	8 0.01
	10 0.01
	12 0.01
	14 0.01
};
\addlegendentry{Polytree}
\addplot [black,mark=square]
table {%
    4 0.0827369747945837
    5 0.0974351989224048
    6 0.0525462494103399
    7 0.0511574819785494
    8 0.0193659417261228
    9 0.0174203018535359
    10 0.00869335343163224
};
\addlegendentry{Trees}
\end{axis}
\draw ({$(current bounding box.south west)!-0.2!(current bounding box.south east)$}|-{$(current bounding box.south west)!0!(current bounding box.north west)$}) node[
scale=0.6,
anchor=base,
text=black,
rotate=0.0
]{.};
\end{tikzpicture}
\caption{Execution times (left) and bounds size (right) for experiments.}\label{fig:time}
\end{figure}
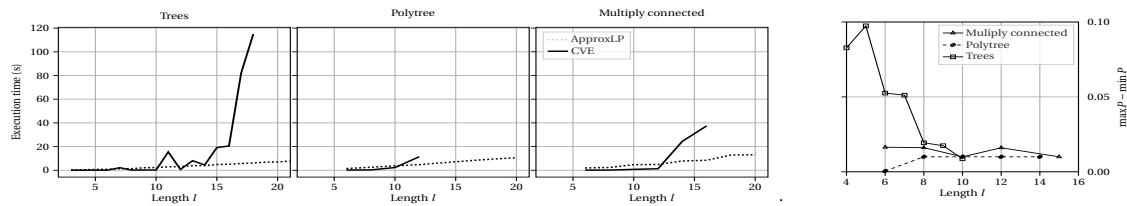

A credal version of \emph{variable elimination} (CVE) is used to compute exact inferences in small networks. Figure \ref{fig:time} (right) depicts the average size of the causal bounds computed with this method. Interestingly, relatively small ($<0.1$), and hence informative, interval sizes are obtained. For larger models we use the \emph{ApproxLP} algorithm for approximate inference in general CNs \citep{approxlp}. Figure \ref{fig:time} (three plots on the left) shows the average execution times of the two methods for different values of $l$. CVE cannot handle large models due time (a timeout of five minutes is set) and space limits. For the sake of readability, the x-axis of these plots ends at $l=20$, but ApproxLP allows to query larger models: with a one-minute timeout, the length of the largest model that can be computed is $l=141$ for trees, $l=92$ for polytrees and $l=38$ for multiply connected models. We also compared the exact intervals returned by CVE with the approximation obtained with ApproxLP. Notably, the average RMSE is low, being $0.61\%$ for trees, $0.03\%$ for polytrees and $0.026\%$ for multiply connected models. The tests have been performed by means of a Java library implementing all the techniques discussed in the paper.\footnote{See \href{http://github.com/idsia/credici}{github.com/idsia/credici} and \href{http://github.com/IDSIA-papers/2020-PGM-structural}{github.com/IDSIA-papers/2020-PGM-structural}.}

\section{Conclusions}\label{sec:conc}
The present work proposes a fully automatic, numeric, approach to causal inference that is alternative to analytical avenues such as do- or $\sigma$-calculus and their specializations, and that is natively capable to solve unidentifiable problems too. It shows that we can take data and an incompletely specified SCM, and turn them, exactly, into a credal network. Standard algorithms for credal nets will then deliver us causal inference up to the top level of \emph{Pearl's causal hierarchy} \citep{bareinboim2020a}: counterfactuals. This holds true even under the extreme condition where the SCM is provided as a bare causal graph, with no information at all about the structural equations---including the cardinality of the $U$ variables (see Remark~\ref{rem:noinfo}). Our algorithms to efficiently convert SCMs into credal nets apply to a wide class of models, which we call quasi-Markovian. It seems possible that these ideas can be extended to more general models and also to the continuous case. The relation with credal networks is, however, already general. In this sense, an SCM is (solvable by) a credal network.

{\small
\bibliography{biblio}}

\begin{thebibliography}{21}
\providecommand{\natexlab}[1]{#1}
\providecommand{\url}[1]{\texttt{#1}}
\expandafter\ifx\csname urlstyle\endcsname\relax
  \providecommand{\doi}[1]{doi: #1}\else
  \providecommand{\doi}{doi: \begingroup \urlstyle{rm}\Url}\fi

\bibitem[Antonucci et~al.(2015)Antonucci, de~Campos, Huber, and
  Zaffalon]{approxlp}
A.~Antonucci, C.~P. de~Campos, D.~Huber, and M.~Zaffalon.
\newblock Approximate credal network updating by linear programming with
  applications to decision making.
\newblock \emph{International Journal of Approximate Reasoning}, 58:\penalty0
  25--38, 2015.

\bibitem[Asghar(2016)]{asghar2016automatic}
N.~Asghar.
\newblock Automatic extraction of causal relations from natural language texts:
  a comprehensive survey.
\newblock \emph{CoRR}, abs/1605.07895, 2016.
\newblock URL \url{http://arxiv.org/abs/1605.07895}.

\bibitem[Balke and Pearl(1994{\natexlab{a}})]{balke1994cf}
A.~Balke and J.~Pearl.
\newblock Probabilistic evaluation of counterfactual queries.
\newblock In B.~Hayes{-}Roth and R.~E. Korf, editors, \emph{Proceedings of
  AAAI/IAAI 1994}, pages 230--237. {AAAI} Press/The {MIT} Press,
  1994{\natexlab{a}}.

\bibitem[Balke and Pearl(1994{\natexlab{b}})]{balke1994counterfactual}
A.~Balke and J.~Pearl.
\newblock Counterfactual probabilities: Computational methods, bounds and
  applications.
\newblock In R.~L. de~M{\'{a}}ntaras and D.~Poole, editors, \emph{Proceedings
  of {UAI} '94}, pages 46--54. Morgan Kaufmann, 1994{\natexlab{b}}.

\bibitem[Balke and Pearl(1997)]{balke1997bounds}
A.~Balke and J.~Pearl.
\newblock Bounds on treatment effects from studies with imperfect compliance.
\newblock \emph{Journal of the American Statistical Association}, 92\penalty0
  (439):\penalty0 1171--1176, 1997.

\bibitem[Bareinboim et~al.(to appear)Bareinboim, Correa, Ibeling, and
  Icard]{bareinboim2020a}
E.~Bareinboim, J.~D. Correa, D.~Ibeling, and T.~Icard.
\newblock On {Pearl'}s hierarchy and the foundations of causal inference.
\newblock In \emph{ACM Special Volume in Honor of Judea Pearl (provisional
  title)}. ACM, to appear.

\bibitem[Correa and Bareinboim(2020)]{correa2020calculus}
J.~D. Correa and E.~Bareinboim.
\newblock A calculus for stochastic interventions: causal effect identification
  and surrogate experiments.
\newblock In \emph{Proceedings of {AAAI} 2020}, pages 10093--10100. {AAAI}
  Press, 2020.

\bibitem[Cozman(2000)]{cozman2000credal}
F.~G. Cozman.
\newblock Credal networks.
\newblock \emph{Artificial intelligence}, 120\penalty0 (2):\penalty0 199--233,
  2000.

\bibitem[Fagiuoli and Zaffalon(1998)]{2u}
E.~Fagiuoli and M.~Zaffalon.
\newblock {2U}: {A}n exact interval propagation algorithm for polytrees with
  binary variables.
\newblock \emph{Artificial Intelligence}, 106\penalty0 (1):\penalty0 77--107,
  1998.

\bibitem[Hume(1739)]{hume}
D.~Hume.
\newblock \emph{A Treatise of Human Nature}.
\newblock Oxford University Press, 1739.

\bibitem[Kang and Tian(2006)]{kang2012inequality}
C.~Kang and J.~Tian.
\newblock Inequality constraints in causal models with hidden variables.
\newblock In \emph{Proceedings of {UAI}~'06}. {AUAI} Press, 2006.

\bibitem[Koller and Friedman(2009)]{koller2009}
D.~Koller and N.~Friedman.
\newblock \emph{Probabilistic Graphical Models: Principles and Techniques}.
\newblock MIT, 2009.

\bibitem[Mau{\'a} et~al.(2014)Mau{\'a}, De~Campos, Benavoli, and
  Antonucci]{maua2014probabilistic}
D.~D. Mau{\'a}, C.~P. De~Campos, A.~Benavoli, and A.~Antonucci.
\newblock Probabilistic inference in credal networks: new complexity results.
\newblock \emph{Journal of Artificial Intelligence Research}, 50:\penalty0
  603--637, 2014.

\bibitem[Pearl(1995)]{pearl1995causal}
J.~Pearl.
\newblock Causal diagrams for empirical research.
\newblock \emph{Biometrika}, 82\penalty0 (4):\penalty0 669--688, 1995.

\bibitem[Pearl(2009)]{pearl2009causality}
J.~Pearl.
\newblock \emph{Causality}.
\newblock {C}ambridge {U}niversity {P}ress, 2009.

\bibitem[Pearl(2010)]{pearl2012measurement}
J.~Pearl.
\newblock On measurement bias in causal inference.
\newblock In P.~Gr{\"{u}}nwald and P.~Spirtes, editors, \emph{Proceedings of
  {UAI} 2010}, pages 425--432. {AUAI} Press, 2010.

\bibitem[Sachs et~al.(2020)Sachs, Gabriel, and S{\"{o}}lander]{sachs2020}
M.~C. Sachs, E.~E. Gabriel, and A.~S{\"{o}}lander.
\newblock Symbolic computation of tight causal bounds.
\newblock \emph{CoRR}, abs/2003.10702, 2020.
\newblock URL \url{http://arxiv.org/abs/2003.10702}.

\bibitem[Shachter(1986)]{shachter1986}
R.~D. Shachter.
\newblock Evaluating influence diagrams.
\newblock \emph{Operations research}, 34\penalty0 (6):\penalty0 871--882, 1986.

\bibitem[Tian and Pearl(2002{\natexlab{a}})]{tian2002b}
J.~Tian and J.~Pearl.
\newblock On the testable implications of causal models with hidden variables.
\newblock In A.~Darwiche and N.~Friedman, editors, \emph{Proceedings of
  {UAI}~'02}, pages 519--527. Morgan Kaufmann, 2002{\natexlab{a}}.

\bibitem[Tian and Pearl(2002{\natexlab{b}})]{tian2002general}
J.~Tian and J.~Pearl.
\newblock A general identification condition for causal effects.
\newblock In R.~Dechter, M.~J. Kearns, and R.~S. Sutton, editors,
  \emph{Proceedings of AAAI/IAAI 2002}, pages 567--573. {AAAI} Press/The {MIT}
  Press, 2002{\natexlab{b}}.

\bibitem[Wilkins(2014)]{wilkins2014practical}
D.~E. Wilkins.
\newblock \emph{Practical Planning: Extending the Classical AI Planning
  Paradigm}.
\newblock Elsevier, 2014.

\end{thebibliography}

\ifarxiv
\appendix
\section{Proofs}\label{sec:proofs}

\begin{proof}[{\bf of Theorem \ref{th:markov}}]
Let us first note that in a Markovian PSCM $(M,P)$, for each $X\in\bm{X}$, $x \in \Omega_X$ and $\underline{\mathrm{pa}}(X)\in\Omega_{\underline{\mathrm{Pa}}(X)}$, we have:
\begin{equation}\label{eq:markovcond}
P(x|\underline{\mathrm{pa}}(X)) 
=\sum_{u\in\Omega_U} P(x|u,\underline{\mathrm{pa}}(X)) \cdot P(u|\underline{\mathrm{pa}}(X))
=\sum_{u\in\Omega_U} P(x|u,\underline{\mathrm{pa}}(X)) \cdot P(u)\,,
\end{equation}
where $U\in\bm{U}$ is the, unique because of Markovianity, exogenous parent of $X$ and, as usual, $\underline{\mathrm{Pa}}(X)$ are the other, endogenous, parents of $X$. The first derivation in Equation \eqref{eq:markovcond} follows from total probability theorem, while the second is because of the \emph{d-separation} between $U$, which is a root node of $\mathcal{G}_M$, and the other parents of its child $X$ \citep{koller2009}.

To prove the theorem, first check the inclusion $\mathcal{K}_{M,\tilde{P}} \subseteq \{K(U)\}_{U\in\bm{U}}$. Take $\{P(U)\}_{U\in\bm{U}} \in \mathcal{K}_{M,\tilde{P}}$. The corresponding PSCM, based on $M$, satisfies Equation \eqref{eq:markovcond} with $P(x|\underline{\mathrm{pa}}(X))=\tilde{P}(x|\underline{\mathrm{pa}}(X))$
 because of the definition of $\mathcal{K}_{M,\tilde{P}}$. Thus:
\begin{equation}\label{eq:markovcond2}
\sum_{u\in\Omega_U} P(x|u,\underline{\mathrm{pa}}(X)) \cdot P(u)=\tilde{P}(x|\underline{\mathrm{pa}}(X))\,.
\end{equation}
Conditional probabilities $P(x|u,\underline{\mathrm{pa}}(X))$ 
in Equation \eqref{eq:markovcond2} are from a (degenerate) CPT  based on the SE $f_X\in M$. We consequently rewrite the equation as:
\begin{equation}\label{eq:markovcond3}
    \sum_{u\in\Omega_U} \llbracket f_X(u,\underline{\mathrm{pa}}(X)) =x \rrbracket \cdot P(u)=\tilde{P}(x|\underline{\mathrm{pa}}(X))\,.
\end{equation}
Equation \eqref{eq:markovcond3} corresponds to Equation \eqref{eq:constraint} (i.e., line 5 of Algorithm \ref{alg:simple}) when $\underline{\mathrm{Pa}}(X)$ is empty and Equation \eqref{eq:constraint_parents} (i.e., line 7 of Algorithm \ref{alg:simple}) otherwise. This proves $P(U)\in K(U)$ for each $U\in\bm{U}$ and hence $\mathcal{K}_{M,\tilde{P}} \subseteq \{K(U)\}_{U\in\bm{U}}$.
Vice versa, to prove $\{K(U)\}_{U\in\bm{U}}\subseteq  \mathcal{K}_{M,\tilde{P}}$ let us take a PMF $P(U)\in K(U)$ for each $U\in\bm{U}$. This induces a (Markovian) PSCM based on $M$ that should satisfy Equation \eqref{eq:markovcond}, but also Equation \eqref{eq:markovcond2} because of lines 5 and 7 of Algorithm \ref{alg:simple}. But this proves the consistency of $P$ with $\tilde{P}$, hence $\{P(U)\}_{U\in\bm{U}} \subseteq \mathcal{K}_{M,\tilde{P}}$ and finally the thesis.
\end{proof}

\begin{lemma}\label{lm:qm}
In a quasi-Markovian PSCM, for each $U\in\bm{U}$, let $\{X_U^k\}_{k=1}^{n_U}$ denote the children of $U$, with the index $k$ sorting them according to a topological order.
For each $x_U^k\in\Omega_{X_U^k}$ and $\underline{\mathrm{pa}}(X_U^k) \in \Omega_{\underline{\mathrm{Pa}}(X_U^k)}$, with $k=1,\ldots,n_U$, we have:
\begin{equation}\label{eq:qmstar}
    \sum_{u\in\Omega_U}P(u) \prod_{k=1}^{n_U} P(x_U^k|\underline{\mathrm{pa}}(X_U^k),u)
    = \prod_{k=1}^{n_U}  P(x_U^{k}|x_U^1,\ldots,x_U^{k-1},\underline{\mathrm{pa}}(X_U^1),\ldots,\underline{\mathrm{pa}}(X_U^k))\,.
\end{equation}

\begin{proof}
Equation \eqref{eq:qmfact} gives a factorization of the joint PMF of the quasi-Markovian PSCM corresponding to that of a BN based on the directed acyclic graph $\mathcal{G}_M$. \emph{Arc reversal} \citep{shachter1986} allows to express the joint PMF of a BN as that of a second BN based on a different directed graph where the orientation of one or more arcs is changed. In our case we want to \emph{reverse} the arcs from $U$ to $X_U^k$, for each $k=1,\ldots,n_U$. To preserve the joint PMF after the reversal of $U\to X_U^k$ we should: (i) add the parents of $U$ to the parents of $X_U^k$, if not already present, and, (ii) add the parents of $X_U^k$ to the parents of $U$, again, if not already present. Although in principle these operation might add cycles to the directed graph, as stated in the proof of Theorem 3 of \citet{shachter1986}, this is not the case if we follow a topological order like the one associated to the index $k$.

To describe the changes induced by these operations to $\mathcal{G}_M$, let us set $\mathcal{G}^0$ equal to $\mathcal{G}_M$, but without the arcs we want to remove and, for each $k=1,\ldots,n_U$, denote as $\mathcal{G}^k$ the causal diagram after the $k$-th reversal, performed together with the above parent augmentations. We similarly denote as $\mathrm{Pa}_k(Z)$ the parents of a generic variable $Z$ in 
$\mathcal{G}^k$. E.g., by our definition of $\mathcal{G}_0$, $\mathrm{Pa}_0(X_U^k)=\underline{\mathrm{Pa}}
(X_U^k)$, for each $k=1,\ldots,n_U$. It is easy to check that, after the first reversal: $\mathrm{Pa}_1(U):=X_U^1\cup\mathrm{Pa}_0(X_U^{1})$, and as $\mathrm{Pa}_0(U)=\emptyset$ (i.e., $U$ is originally a root), $\mathrm{Pa}_1({X_U^1}):=\mathrm{Pa}_0({X_U^1})$. All the other sets of parents are unchanged. After $k$ reversals we have instead:
\begin{eqnarray}\label{eq:pau}
    \mathrm{Pa}_{k}(U)&:=& \bigcup_{i=1}^k \left[X_U^i\cup\mathrm{Pa}_0({X_U^i})\right]\,,\\
    \mathrm{Pa}_k(X_U^k)&:=&
    \left[
    \bigcup_{i=1}^{k-1} X_U^{i}
    \right]\cup
    \left[
    \bigcup_{i=1}^{k} \mathrm{Pa}_0({X_U^i})
    \right]\,.\label{eq:pax}
\end{eqnarray}
The validity of these two equations for each $k=1,\ldots,n_U$ can be proved by induction. First note that they are satisfied for $k=1$. After that, assume Equations \eqref{eq:pau} and \eqref{eq:pax} valid for $k=j-1$ and prove that these also hold for $k=j$. After the $j$-th reversal, the parents of $U$ can be expressed as:
\begin{equation}\label{eq:recursion}
\mathrm{Pa}_j(U)=\mathrm{Pa}_{j-1}(U) \cup X_U^j \cup \mathrm{Pa}_{j-1}(X_U^j) = \cup_{i=1}^{j-1} \left[ X_U^j \cup \mathrm{Pa}_0(X_U^i) \right]\cup X_U^j \cup \mathrm{Pa}_0(X_U^j)\,,
\end{equation}
where in the first derivation we added to the parents of $U$, as they are before the reversal, $X_U^j$ and its parents (again before the reversal). The second step follows from the validity of the Equation \eqref{eq:pau}  for $k=j-1$ and from the general fact $\mathrm{Pa}_l(X_U^j)=\mathrm{Pa}_{0}(X_U^j)$ for $l<j$ (i.e., the parents of $X_U^j$ remain unchanged before the $j$-th reversal). It is easy to check that the rightmost-hand side of Equation \eqref{eq:recursion} corresponds to the right-hand side of Equation \eqref{eq:pau} for $k=j$. Similarly, for $X_U^j$:
\begin{equation}\label{eq:recursion2}
\mathrm{Pa}_j(X_U^j)=\mathrm{Pa}_{j-1}(X_U^j) \cup \mathrm{Pa}_{j-1}(U) 
=
\mathrm{Pa}_{0}(X_U^j) \cup
\bigcup_{i=1}^{j-1} \left[X_U^i\cup\mathrm{Pa}_0({X_U^i})\right]\,,
\end{equation}
where in the first derivation we added to the parents of $X_U^j$, as they are before the reversal, the parents of $U$ (again before the reversal). The second step follows from the above discussed fact that the parents of $X_U^j$ are unchanged before the $j$-th reversal and the validity of Equation \eqref{eq:pax} for $k=j-1$. It is easy to check that the rightmost-hand side of Equation \eqref{eq:recursion2} corresponds to the right-hand side of Equation \eqref{eq:pax} for $k=j$. Overall, we have Equations \eqref{eq:pau} and \eqref{eq:pax} valid for each $k=1,\ldots,n_U$. An example of these reversals is in Figure \ref{fig:rev}.

Note that after the $n_U$ reversals, $U$ is a barren node as: (i) all the $n_U$ original children of $U$ are now its parents; (ii) for each $k=1,\ldots,n_U$, $U$ does not belong to the new parents of $X_U^k$ in Equation \eqref{eq:pax}. After the last reversal, the joint PMF associated with the quasi-Markovian PSCM is still factorizing as a BN, but based on a different directed acyclic graph. The new CPTs are those associated with $U$ and its children, while all the other ones remain unchanged. The parents of $U$ are now as in Equation \eqref{eq:pau}, to be considered for $k=n_U$. For each $k=1,\ldots,n_U$, the parents of $X_U^k$ are provided instead by Equation \eqref{eq:pax} (remember that after the $k$-th reversal, the parents of $X_U^k$ will not be further modified).
As the new ``reversed'' joint PMF with the original one in Equation \eqref{eq:qmfact} coincide, we can simplify from both sides all the probabilities associated to the CPTs that are not associated with $U$ and its children and obtain:
\begin{align}\label{eq:qmstar2}
P(u) \prod_{k=1}^{n_U} P(x_U^k|\underline{\mathrm{pa}}(X_U^k),u)=\\ \nonumber
P(u|x_U^1,\ldots,x_U,\underline{\mathrm{pa}}(X_U^1),\ldots,\underline{\mathrm{pa}}(X_U^k))
\prod_{k=1}^{n_U}
P(x_U^{k}|x_U^1,\ldots,x_U^{k-1},\underline{\mathrm{pa}}(X_U^1),\ldots,\underline{\mathrm{pa}}(X_U^k))\,.
\end{align}
Finally, Equation \eqref{eq:qmstar} follows from Equation \eqref{eq:qmstar2} by taking the sum over $U$ on both sides and noticing that, as $U$ is a barren node, it only appears in its own CPT.
\end{proof}
\end{lemma}

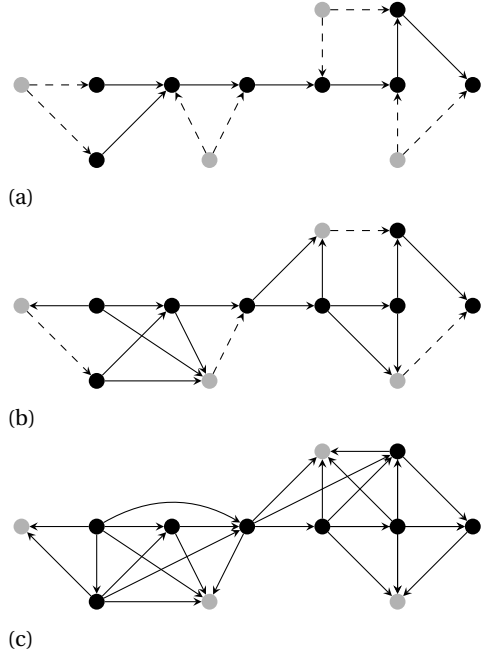
\begin{figure}\centering
\begin{subfigure}[]{}
\begin{tikzpicture}[]
\node[dot] (x1)  at (0,0) {};
\node[dot] (x2)  at (0,-1) {};
\node[dot] (x3)  at (1,0) {};
\node[dot] (x4)  at (2,0) {};
\node[dot] (x5)  at (3,0) {};
\node[dot] (x6)  at (4,0) {};
\node[dot] (x7)  at (4,1) {};
\node[dot] (x8)  at (5,0) {};
\node[dot,color=black!30] (u1)  at (-1,0) {};
\node[dot,color=black!30] (u2)  at (1.5,-1) {};
\node[dot,color=black!30] (u3)  at (3,1) {};
\node[dot,color=black!30] (u4)  at (4,-1) {};
\draw[a2] (x1) -- (x3);
\draw[a2] (x2) -- (x3);
\draw[a2] (x3) -- (x4);
\draw[a2] (x4) -- (x5);
\draw[a2] (x5) -- (x6);
\draw[a2] (x6) -- (x7);
\draw[a2] (x7) -- (x8);
\draw[a] (u1) -- (x1);
\draw[a] (u1) -- (x2);
\draw[a] (u2) -- (x3);
\draw[a] (u2) -- (x4);
\draw[a] (u3) -- (x5);
\draw[a] (u3) -- (x7);
\draw[a] (u4) -- (x6);
\draw[a] (u4) -- (x8);
\end{tikzpicture}
\end{subfigure}

\begin{subfigure}[]{}
    \begin{tikzpicture}[]
    \node[dot] (x1)  at (0,0) {};
    \node[dot] (x2)  at (0,-1) {};
    \node[dot] (x3)  at (1,0) {};
    \node[dot] (x4)  at (2,0) {};
    \node[dot] (x5)  at (3,0) {};
    \node[dot] (x6)  at (4,0) {};
    \node[dot] (x7)  at (4,1) {};
    \node[dot] (x8)  at (5,0) {};
    \node[dot,color=black!30] (u1)  at (-1,0) {};
    \node[dot,color=black!30] (u2)  at (1.5,-1) {};
    \node[dot,color=black!30] (u3)  at (3,1) {};
    \node[dot,color=black!30] (u4)  at (4,-1) {};
    \draw[a2] (x1) -- (x3);
    \draw[a2] (x2) -- (x3);
    \draw[a2] (x1) -- (u2);
    \draw[a2] (x2) -- (u2);
    \draw[a2] (x4) -- (u3);
    \draw[a2] (x6) -- (u4);
    \draw[a2] (x3) -- (x4);
    \draw[a2] (x4) -- (x5);
    \draw[a2] (x5) -- (x6);
    \draw[a2] (x6) -- (x7);
    \draw[a2] (x7) -- (x8);
    \draw[a2] (x1) -- (u1);
    \draw[a] (u1) -- (x2);
    \draw[a2] (x3) -- (u2);
    \draw[a] (u2) -- (x4);
    \draw[a2] (x5) -- (u3);
    \draw[a] (u3) -- (x7);
    \draw[a2] (x5) -- (u4);
    \draw[a] (u4) -- (x8);
    \end{tikzpicture}
\end{subfigure}

\begin{subfigure}[]{}
    \begin{tikzpicture}[]
    \node[dot] (x1)  at (0,0) {};
    \node[dot] (x2)  at (0,-1) {};
    \node[dot] (x3)  at (1,0) {};
    \node[dot] (x4)  at (2,0) {};
    \node[dot] (x5)  at (3,0) {};
    \node[dot] (x6)  at (4,0) {};
    \node[dot] (x7)  at (4,1) {};
    \node[dot] (x8)  at (5,0) {};
    \node[dot,color=black!30] (u1)  at (-1,0) {};
    \node[dot,color=black!30] (u2)  at (1.5,-1) {};
    \node[dot,color=black!30] (u3)  at (3,1) {};
    \node[dot,color=black!30] (u4)  at (4,-1) {};
    \draw[a2] (x1) -- (x3);
    \draw[a2] (x2) -- (x3);
    \draw[a2] (x1) -- (u2);
    \draw[a2] (x2) -- (u2);
    \draw[a2] (x4) -- (u3);
    \draw[a2] (x6) -- (u4);
    \draw[a2] (x3) -- (x4);
    \draw[a2] (x4) -- (x5);
    \draw[a2] (x5) -- (x6);
    \draw[a2] (x6) -- (x7);
    \draw[a2] (x7) -- (x8);
    \draw[a2] (x1) -- (u1);
    \draw[a2] (x2) -- (u1);
    \draw[a2] (x1) -- (x2);    
    \draw[a2] (x2) -- (x4);    
    \draw[bend left,a2] (x1) edge (x4);
    \draw[a2] (x3) -- (u2);
    \draw[a2] (x4) -- (u2);
    \draw[a2] (x5) -- (u3);
    \draw[a2] (x6) -- (u3);
    \draw[a2] (x5) -- (u4);
    \draw[a2] (x7) -- (u3);
    \draw[a2] (x7) -- (u4);
    \draw[a2] (x4) edge (x7);
    \draw[a2] (x5) edge (x7);
    \draw[a2] (x5) edge (x8);
    \draw[a2] (x6) edge (x8);
    \draw[a2] (x8) -- (u4);
    \end{tikzpicture}
\end{subfigure}
\caption{Arc reversals in a quasi-Markovian SCM (exogenous nodes are in gray).\label{fig:rev}}
\end{figure}

\begin{proof}[{\bf of Theorem \ref{th:quasi}}]
The proof is analogous to that of Theorem \ref{th:markov}, with Equation \eqref{eq:qmstar} from Lemma \ref{lm:qm} playing the role of Equation \eqref{eq:markovcond}.

To prove the theorem, first check the inclusion $\mathcal{K}_{M,\tilde{P}} \subseteq \{K(U)\}_{U\in\bm{U}}$. Take $\{P(U)\}_{U\in\bm{U}} \in \mathcal{K}_{M,\tilde{P}}$. The corresponding quasi-Markovian PSCM, based on $M$, satisfies Equation \eqref{eq:qmstar} with the product of conditional probabilities in the right-hand side as in the empirical case because of the definition of $\mathcal{K}_{M,\tilde{P}}$. Thus:
\begin{equation}\label{eq:qmarkovcond2}
    \sum_{u\in\Omega_U}P(u) \prod_{k=1}^{n_U} P(x_U^k|\underline{\mathrm{pa}}(X_U^k),u)
= \prod_{k=1}^{n_U}  \tilde{P}(x_U^{k}|x_U^1,\ldots,x_U^{k-1},\underline{\mathrm{pa}}(X_U^1),\ldots,\underline{\mathrm{pa}}(X_U^k))\,.
\end{equation}
The conditional probabilities in the left-hand side of Equation \eqref{eq:qmarkovcond2} are from (degenerate) CPTs based on the SEs of the children of $U$. We consequently rewrite the equation as:
\begin{equation}\label{eq:qmarkovcond3}
    \sum_{u\in\Omega_U}P(u) \prod_{k=1}^{n_U} \llbracket f_{X_U^k}(\underline{\mathrm{pa}}(X_U^k),u) = x_U^k \rrbracket
= \prod_{k=1}^{n_U}  \tilde{P}(x_U^{k}|x_U^1,\ldots,x_U^{k-1},\underline{\mathrm{pa}}(X_U^1),\ldots,\underline{\mathrm{pa}}(X_U^k))\,.
\end{equation}
Equation \eqref{eq:qmarkovcond3} corresponds to the linear constraint specification in line 7 of Algorithm \ref{alg:quasi}.
This proves $P(U)\in K(U)$ for each $U\in\bm{U}$ and hence $\mathcal{K}_{M,\tilde{P}} \subseteq \{K(U)\}_{U\in\bm{U}}$.

Vice versa, to prove $\{K(U)\}_{U\in\bm{U}}\subseteq  \mathcal{K}_{M,\tilde{P}}$ let us take a PMF $P(U)\in K(U)$ for each $U\in\bm{U}$. This induces a quasi-Markovian PSCM based on $M$ that should satisfy Equation \eqref{eq:qmstar}, but also Equation \eqref{eq:qmarkovcond2} because of line 7 of Algorithm \ref{alg:quasi}. But this proves the consistency of $P$ with $\tilde{P}$, hence $\{P(U)\}_{U\in\bm{U}} \subseteq \mathcal{K}_{M,\tilde{P}}$ and finally the thesis.
\end{proof}
\fi
\end{document}